\documentclass[runningheads]{llncs}
\usepackage[T1]{fontenc}
\usepackage{lipsum}
\usepackage{amsmath}
\usepackage{amssymb}
\usepackage{url}
\usepackage{doi}
\usepackage{multirow}
\usepackage{booktabs}
\usepackage{xcolor}
\usepackage[skip=2pt]{caption}
\usepackage{tabularx}
\usepackage{longtable}
\usepackage{graphicx}
\usepackage{subcaption} 

\makeatletter
\def\blfootnote{\gdef\@thefnmark{}\@footnotetext}
\makeatother
\usepackage{graphicx}
\usepackage{placeins}
\usepackage[ruled,vlined,linesnumbered]{algorithm2e}

\begin{document}

\title{RAC: Retrieval-Augmented Clarification for Faithful Conversational Search}
%
%
\author{Ahmed Rayane Kebir\inst{1,2}\orcidID{0009-0009-2512-832X} \and
Vincent Guigue\inst{3}\orcidID{0000-0002-1450-5566} \and
Lynda Said Lhadj\inst{2}\orcidID{0009-0005-3850-9229} \and
Laure Soulier\inst{1}\orcidID{0000-0001-9827-7400}}

\institute{
Sorbonne Université, CNRS, ISIR, F-75005 Paris, France  \\  %
\and Ecole nationale Supérieure d’Informatique (ESI), Algeria \\
\and AgroParisTech, UMR MIA-PS, Palaiseau, France
}

\blfootnote{%
  \makebox[\textwidth][c]{%
    \begin{minipage}{1.4\textwidth} 
    \footnotesize
  This is the author's version of the work. It is posted here for your personal use. The definitive version is published in:\\
\textit{Proc of the 48th European Conference on Information Retrieval (ECIR '26), 29 March--2 April, 2026, Delft, Netherlands}
    \end{minipage}
  }%
}

\authorrunning{A-R Kebir et al.}
\titlerunning{Retrieval-Augmented Clarification for Faithful CS}
%
%
\maketitle              
\begin{abstract}
\sloppy
Clarification questions help conversational search systems resolve ambiguous or underspecified user queries. While prior work has focused on fluency and alignment with user intent, especially through facet extraction, much less attention has been paid to grounding clarifications in the underlying corpus. Without such grounding, systems risk asking questions that cannot be answered from the available documents.
We introduce RAC (\textbf{R}etrieval-\textbf{A}ugmented \textbf{C}larification), a framework for generating corpus-faithful clarification questions. 
After comparing several indexing strategies for retrieval,
we fine-tune a large language model to make optimal use of research context and to encourage the generation of evidence-based question.
We then apply contrastive preference optimization to favor questions supported by retrieved passages over ungrounded alternatives.
Evaluated on four benchmarks, RAC demonstrate significant improvements over baselines. 
In addition to LLM-as-Judge assessments, we introduce novel metrics derived from NLI and data-to-text to assess how well questions are anchored in the context, and we demonstrate that our approach consistently enhances faithfulness.

\end{abstract}
\section{Introduction}
In open-domain information-seeking tasks, user queries are often short, ambiguous, or under-specified. Such characteristics make it difficult for traditional search systems to accurately capture user intent, as they typically provide only a ranked list of documents or passages without engaging in clarifying interactions~\cite{radlinski2017theoretical}.
Recent work has explored generating clarifying questions that are relevant, diverse, and human-plausible \cite{erbacher2024paqa,siro_agent-cq_2024,wang2023zero}. However, little attention has been given to whether these questions are grounded in the document corpus, even though unsupported clarifications may mislead users and harm retrieval effectiveness \cite{krasakis2024corpus,mass-etal-2022-conversational}.

Early approaches to clarifying question generation in conversational search largely relied on facet-based methods. These methods extracted candidate facets from the document collection to produce clarifying questions via templates or sequence-to-sequence models \cite{aliannejadi2021building,aliannejadi2019asking}. While this offered a basic form of corpus grounding, the reliance on coarse-grained facets proved reductive.

The advent of large language models (LLMs) enabled more fluent generation, with systems either conditioning on extracted facets to produce natural clarifications or directly deriving facets from queries before turning them into questions. Yet the task remains split into two stages—facet identification and question generation—creating bottlenecks in facet extraction and risks of hallucination when clarifications introduce content unsupported by the corpus~\cite{sekulic_towards_2021,siro_agent-cq_2024}.

In this work, we build on the retrieval-augmented generation (RAG) paradigm ~\cite{lewis2021retrievalaugmentedgenerationknowledgeintensivenlp} to ground clarifications directly in the corpus, focusing on answers supported by the documents. Facet extraction is performed implicitly by supplying the top-$k$ retrieved passages to the LLM, which then generates the clarifying question.
The first contribution of this article is to propose a fine-tuning of conditional clarification generation, which greatly improves the quality of the questions.
To further mitigate entity-level hallucinations, we also introduce a faithfulness reinforcement mechanism that steers the model to rely on the retrieved inputs rather than its internal knowledge, following the approach of~\cite{duong2025scope}.

\begin{figure}[t]
    \centering
    \includegraphics[width=1\linewidth]{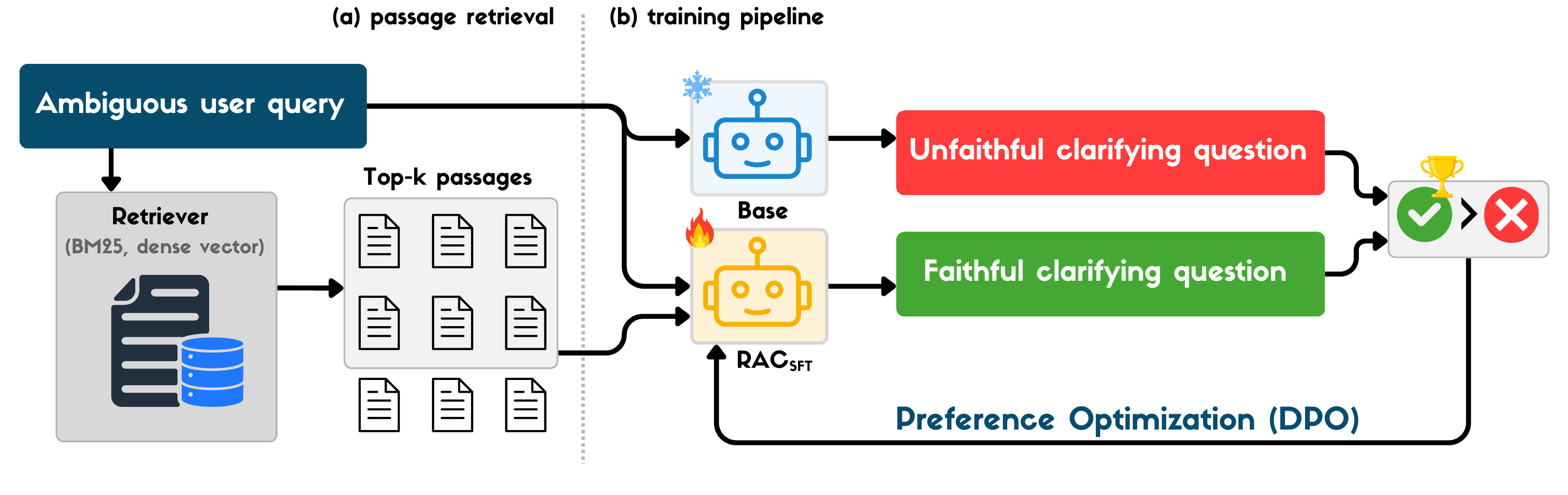}

\caption{Overview of RAC. Given an ambiguous user query, the system first retrieves the top-$k$ passages ((a) passage retrieval). A mixture of the fine-tuned model and the base model is then used to generate unfaithful clarifying questions. Both faithful and unfaithful clarifying questions are subsequently leveraged for preference optimization via the DPO algorithm ((b) training pipeline). During inference, the trained model directly generates faithful clarifying questions.}
    \label{fig:rac-example-pipeline}
\end{figure}

\noindent
Thus, we aim to address the following research questions:
\begin{itemize}
\item[] \textbf{RQ1.} How can relevant passages be selected from the corpus, and how many should be used to optimally guide clarification?
\item[] \textbf{RQ2.} How does conditioning on these relevant passages affect the generation of clarifying questions?
\item[] \textbf{RQ3.} How can the faithfulness of clarifying question generation be improved when conditioned on relevant passages?
\end{itemize}

We introduce RAC, a framework for generating clarifying questions grounded in relevant retrieved passages, and train a large language model to prioritize faithful questions using preference tuning and contrastive learning, as illustrated in Fig.~\ref{fig:rac-example-pipeline}. We validate our approach on conversational search and open-domain Question Answering datasets through automatic metrics and LLM-as-Judge evaluations. Results show that RAC consistently enhances both the quality and faithfulness of clarifying questions, outperforming existing baselines.

\section{Related Work}

\paragraph{\textbf{Query Clarification in Conversational Search.}} 
Asking clarifying questions enables users to actively participate in query disambiguation, with the goal of better capturing their information intent~\cite{aliannejadi2021building,aliannejadi2019asking,erbacher_augmenting_2023,lee2023asking}. Prior work in this area has primarily focused on two tasks: predicting the need for clarification and generating clarifying questions. In this paper, we focus on the latter. Recent studies have increasingly explored large language model based approaches. For instance, Sekulić et al.~\cite{sekulic_towards_2021} conditioned an LLM on specific facets; however, such facets are not always readily available and often require external extraction tools. Siro et al.~\cite{siro_agent-cq_2024} leveraged temperature control and facet information to generate diverse clarifications, while Wang et al.~\cite{wang2023zero} introduced a zero-shot clarifying-question generator using fixed templates and query facets. More recently, Tang et al.~\cite{tang2025clarifying} proposed a prompting strategy grounded in an ambiguity taxonomy to improve handling of ambiguous queries. Although these methods produce plausible and diverse clarifications, they remain prone to hallucination, frequently generating questions about aspects unsupported by the underlying corpus. Additionally, the reliance on explicit facets limits applicability when facets are difficult to extract or unavailable. 

\paragraph{\textbf{Retrieval Augmented Generation.}}
Since the original article~\cite{lewis2021retrievalaugmentedgenerationknowledgeintensivenlp}, several variants have been proposed, first for question answering~\cite{izacard-grave-2021-leveraging} and later for clarification. Early studies primarily examined the role of the retriever in selecting corpus-grounded clarifications among candidate suggestions~\cite{mass-etal-2022-conversational}, whereas more recent work has shifted the focus toward generation~\cite{krasakis2024corpus}, with particular attention to maximizing faithfulness during inference. In addition, \cite{sahay2025ask} demonstrates that the RAG paradigm can be combined with knowledge bases to enhance disambiguation in domain-specific applications. However, these approaches rely on a zero-shot paradigm, whereas we demonstrate the benefit of fine-tuning the generator to better exploit the retrieved passages.

\paragraph{\textbf{Preference Tuning.}}

Reinforcement learning from human feedback was introduced to align LLMs with human preferences~\cite{ouyang2022traininglanguagemodelsfollow}, but reward-model methods were costly and often unstable. More recent techniques such as direct preference optimization (DPO)~\cite{rafailov2023direct} and extensions~\cite{yang2025ipo} have improved efficiency by learning directly from pairwise comparisons. 
Beyond general alignment, generating both faithful and unfaithful baseline sentences allows contrastive learning algorithms to be effectively applied for improving text generation. Such approaches have demonstrated strong performance in tasks such as automatic summarization~\cite{choi2024model} and data-to-text generation~\cite{duong2025scope}.
To be useful, text variants must be generated carefully, and previous work has relied on mixture-of-logits decoding.
Such techniques are directly relevant to conversational search, where clarifying questions must remain faithful to the corpus.

\paragraph{\textbf{Faithfulness Evaluation.}}
\sloppy
Faithfulness measures whether generated text remains consistent with its input. In summarization, state-of-the-art approaches employ entailment-based metrics that leverage NLI models to score the consistency of summaries with source documents (e.g., RoBERTa-based entailment~\cite{liu2019roberta}). These methods provide fine-grained judgments of factual alignment on a continuous scale. In data-to-text generation, metrics such as PARENT~\cite{dhingra2019handling} evaluate whether candidate outputs faithfully express entities and relations from structured inputs.
By contrast, clarifying question generation has not been systematically assessed for faithfulness. Existing evaluations rely mainly on reference-based metrics (e.g., BLEU, METEOR) or indirect retrieval-based proxies ~\cite{aliannejadi2019asking,Rao2018LearningTA}, which do not directly measure factual consistency with the input context. In this work, we adapt entailment-based and data-grounding approaches from summarization and data-to-text to develop faithfulness evaluations tailored to clarifying question generation.

\section{Methodology}

\begin{figure}[b]
\centering
\includegraphics[width=0.8\textwidth]{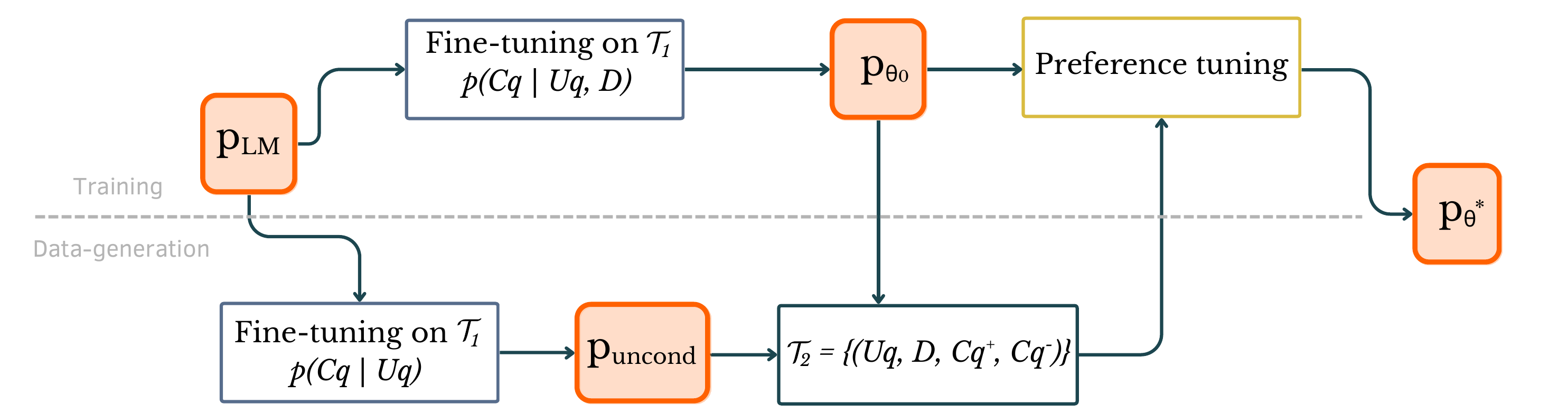}
\caption{Overview of our proposed training pipeline.} \label{fig:pipeline}
\end{figure}

Our RAC framework follows a two-stage training pipeline as illustrated in Fig.~\ref{fig:pipeline}. In the first stage, a large language model $p_{LM}$ is fine-tuned on existing clarification datasets along two axes to generate: (1) factual questions conditioned by user queries and retrieved passages $p_{\theta_0}$ and (2) less factual questions unconditioned by passages $p_{\text{uncond}}$. The two levels of question quality are then passed to a preference learning algorithm (contrastive) that encourages the model to rank faithful, evidence-grounded clarifications higher than unsupported or hallucinated alternatives. 

We formulate the task of generating clarifying questions $Cq$ as a retrieval-augmented generation task. The initial user query $U_q$ enables the retrieval of a set of relevant passages $\mathcal D = \{d_1, \ldots, d_N\}$, which will be used as context for the generation. We assume all queries to be ambiguous, focusing on clarifying question generation rather than clarification need prediction ~\cite{lu2025zero}. Each passage may capture different semantic facets of the query, but we restrict to a single-turn setup, generating one clarifying question targeting the most useful facet.

\subsection{Supervised Clarifying Question Generation}
Retrieval-augmented generation (RAG) has shown that conditioning large language models on retrieved passages improves factual grounding and reduces reliance on parametric memory~\cite{izacard-grave-2021-leveraging,lewis2021retrievalaugmentedgenerationknowledgeintensivenlp}. However, previous work has focused on generating direct zero-shot answers. Our contribution is to propose a fine-tuned model (twice) to better exploit the retrieved passages for the clarification task. To this end, we employ supervised fine-tuning (SFT) as the first stage of training: a large language model is trained to generate clarifying questions $C_q$ conditioned on both the user query $U_q$ and the corresponding retrieved passages $\mathcal D$ (leading to $p_{\theta_0}$). Given a dataset $\mathcal{T}_1$ of query--passage--ground-truth-question tuples $(U_q, \mathcal D, C_q^+)$, the model is optimized with the negative log-likelihood objective:
\begin{equation}
\mathcal{L}_{\text{SFT}}(\theta) = - \mathbb{E}_{\sim \mathcal{T}_1} \left[ \sum_{t=1}^{|C_q^+|} \log p_\theta\!\left(C_{q,t} \,\middle|\, U_q, \mathcal D, C_{q,    <t}\right) \right]
\label{eq:sft_loss}
\end{equation}
Here, each token of the clarifying question $C_{q,t}$ is predicted sequentially, conditioned on the user query, the retrieved passages, and the previously generated tokens (denoted $C_{q,<t}$).

SFT establishes a strong baseline for clarification. By learning to ask questions supported by retrieved passages, the model reduces ambiguity in user intent and provides an evidence-aligned starting point for the subsequent preference based alignment stage. This further improves faithfulness and mitigates hallucinations.

\subsection{Faithfulness Alignment}
\label{subsection:faithfulness}

Although the $p_{\theta_0}$ model is already fine-tuned to generate clarifying questions that are much more relevant than the initial $p_{LM}$ model, one of its main limitations is its tendency to hallucinate: it may generate details that are absent from the retrieved passages $\mathcal D$.

\paragraph{\textbf{Preference tuning.}}
\sloppy
To mitigate this, we introduce a second training stage focused on faithfulness. We augment the training data with pairs of faithful ($C_q^+$) and unfaithful ($C_q^-$) clarifying questions and apply a contrastive learning approach. In particular, we employ DPO~\cite{rafailov2023direct} over a dataset $\mathcal{T}_2 = \{ (U_q, \mathcal D, C_q^+, C_q^-) \}$, where the model is explicitly trained to prefer faithful clarifying questions over unfaithful ones. In DPO, the learning objective (Eq.~\ref{eq:dpo_loss}) aligns a policy model $p_\theta$ with a preference signal, favoring $C_q^+$ over $C_q^-$, given the same input $(U_q, \mathcal D)$, as defined below:

\begingroup
\small
\begin{equation}
\mathcal{L}_{\text{DPO}}(\theta)
= - \mathbb{E}_{\sim \mathcal{T}_2} \Big[
\log \sigma \Big(
\beta \log \frac{p_\theta(C_q^{+} \mid U_q,\mathcal D)}{p_{\theta_0}(C_q^{+} \mid U_q,\mathcal D)}
- \beta \log \frac{p_\theta(C_q^{-} \mid U_q,\mathcal D)}{p_{\theta_0}(C_q^{-} \mid U_q,\mathcal D)}
\Big)
\Big]
\label{eq:dpo_loss}
\end{equation}
\endgroup

\paragraph{\textbf{Unfaithful clarifying questions generation.}}

Preference-based alignment requires faithful–unfaithful question pairs, but manual creation is costly and automatic detection remains difficult. We propose an unsupervised method that simulates unfaithful questions by injecting controlled noise during decoding. Our method adapts the noisy decoding strategy of Duong et al.~\cite{duong2025scope} to the clarification setting. The approach relies on two complementary models:

\noindent
\texttt{Grounded model $p_{\theta_0}$:} obtained by fine-tuning a pretrained base model $p_{LM}$ on half of the dataset $\mathcal{T}_1$. Given a query and retrieved passages $(U_q, \mathcal D)$, it outputs generally faithful clarifying questions $C_q \sim p_{\theta_0}(\cdot \mid U_q, \mathcal D)$, though minor inaccuracies remain.

\noindent
\texttt{Ungrounded model $p_{\text{uncond}}$:} obtained by fine-tuning the same base model but conditioned only on the user query $U_q$, i.e., $C_q \sim p_{uncond}(\cdot \mid U_q)$. It produces fluent and relevant clarifying questions, yet these are not guaranteed to be grounded in the retrieved passages $\mathcal D$.

While $p_{uncond}$ produces overly unconstrained questions and $p_{\theta_0}$ tends to remain faithful, their combination yields plausible but unfaithful clarifying questions (the balance is critical, as highlighted by Duong et al.~\cite{duong2025scope}). Specifically, we decode token-by-token from a mixture distribution (Eq.~\ref{eq:mix_dist}), using stochastic decoding (temperature and top-$k$ sampling) to promote diversity and encourage hallucinated tokens.
\begin{equation}
C_{q,t} \sim (1-\alpha_t)\, p_{\theta_0}(\cdot \mid C_{q,<t}, U_q, \mathcal D) \ + \ \alpha_t \, p_{\text{uncond}}(\cdot \mid C_{q,<t}, U_q),
\label{eq:mix_dist}
\end{equation}
where $\alpha_t \sim \text{Bernoulli}(\alpha)$ controls the injection of ungrounded content. The noise parameter $\alpha \in [0,1]$ determines the faithfulness–fluency trade-off: $\alpha=0$ recovers clarifying questions from $p_{\theta_0}$, whereas $\alpha=1$ generates ungrounded ones from $p_{uncond}$. The resulting questions remain fluent but contain ungrounded spans, yielding both intrinsic errors (contradictions with retrieved passages) and extrinsic hallucinations (additions not inferable from $\mathcal D$). These are used as unfaithful clarifying questions $C_q^-$ in the augmented dataset $\mathcal{T}_2 = { (U_q, \mathcal D, C_q^+, C_q^-) }$, enabling preference optimization for faithfulness alignment.

\subsection{Joint Training Objective}

Supervised fine-tuning and preference optimization address complementary objectives: supervised fine-tuning operates at the token level, teaching the model to produce clarifying questions, while preference optimization encourages it to prefer faithful outputs over unfaithful ones. To leverage both, we propose a combined training objective: $\mathcal{L}_{\text{RAC}}(\theta) = \gamma\cdot\mathcal{L}_{\text{DPO}}(\theta) \ + \ \  (1-\gamma)\cdot \mathcal{L}_{\text{SFT}}(\theta).$

\section{Experimental Setup}


\subsection{Datasets and Evaluation}

\subsubsection{Datasets.} We evaluate RAC on four datasets across conversational search and open-retrieval QA. For search, we use Qulac (derived from TREC Web Track 2009–2012)~\cite{aliannejadi2019asking} and the filtered version of ClariQ proposed by Sekulic et al.~\cite{sekulic_towards_2021}, which maps clarifying questions to facets. For QA, we use PaQa (AmbigNQ with GPT-3 clarifications)~\cite{erbacher2024paqa} and CambigNQ (AmbigNQ queries augmented with human-validated clarifications)~\cite{lee2023asking}.

\paragraph{Adapting Datasets for Retrieval-Augmented Clarification.}
Existing clarification datasets (Qulac, ClariQ) lack passage-level grounding, as their relevance labels are assigned at the document level and not explicitly tied to the clarifying question. To bridge this gap, we derive passage-level supervision through a three-stage pipeline: (i) \textit{Passage Indexing}: we segment Clueweb09-12\footnote{https://lemurproject.org/clueweb09/} into 250-token passages, following TREC CAsT~\cite{owoicho2022trec}, and index them with Pyserini~\cite{lin2021pyserini}; (ii) \textit{Query Rewriting}: for each ambiguous query–clarification pair $(U_q, C_q)$, we generate a facet-specific reformulation $U_q^r$ by incorporating $C_q$ using an LLM, yielding sharper retrieval intents than $U_q$ alone; (iii) \textit{Pseudo-Relevance Retrieval}: we employ BM25~\cite{robertson2009probabilistic} over the passage index to retrieve the top-$k$ passages $\mathcal D$ for $U_q^r$, treating them as pseudo-relevant evidence. This produces training tuples $(U_q, \mathcal D, C_q)$ that support retrieval-conditioned clarification generation.

\subsubsection{Metrics.}

We employ both reference-based and reference-free metrics to evaluate the quality of generated clarifying questions. Reference-based metrics measure similarity to gold questions, while reference-free metrics assess faithfulness to the input query and associated passages. In addition, we use GPT-4 to assess faithfulness, serving as a model-based proxy for human judgment.

\paragraph{Reference-based evaluation.}
We report BLEU~\cite{papineni2002bleu}, ROUGE-L~\cite{lin2004rouge}, METEOR~\cite{banerjee2005meteor}, and BERTScore~\cite{zhang2019bertscore}. BLEU and ROUGE-L capture n-gram and longest common subsequence overlap, respectively, while METEOR accounts for synonym and stem matches. BERTScore computes semantic similarity via contextualized token embeddings, providing a finer-grained assessment of meaning preservation. These metrics are consistent with prior work in clarification question generation and facilitate direct comparison.

\paragraph{Faithfulness evaluation.}We evaluate faithfulness using PARENT Recall (PAR) \cite{dhingra2019handling} and AlignScore (AL) \cite{zha2023alignscore}. PAR, originally proposed for data-to-text generation, computes n-gram recall against both the input and the reference, serving as a proxy for input-groundedness. To apply it to unstructured passages, we adapt the metric by extracting named entities, multi-word noun phrases, and subject–verb–object triples with SpaCy\footnote{https://spacy.io/}, allowing content-level overlap measurement without reliance on structured data. AL is an entailment-based metric built on RoBERTa \cite{liu2019roberta} and trained on multiple NLI datasets. Because clarifying questions are often interrogative and not well-suited for direct entailment evaluation, we convert them into declarative statements by removing question templates, retaining only content-bearing tokens, and filtering query overlaps. This yields hypotheses compatible with AL’s premise–hypothesis structure while preserving the semantic content of the questions.

\subsection{Baselines}
We evaluate RAC against several baselines. First, we include (\texttt{AT-CoT}), the ambiguity taxonomy chain-of-thought prompting baseline of Tang et al.~\cite{tang2025clarifying}, which applies few-shot prompting conditioned only on the query. Following Sekulic et al.~\cite{sekulic_towards_2021}, we use the widely adopted (\texttt{Q-Cond}) fine-tuned model, which generates clarifications from the query alone. To assess the impact of supervision, we compare RAC to a (\texttt{QP-Zero\textsubscript{shot}}) variant conditioned on both query and passages in a zero-shot setting. Finally, on ClariQ, where facet annotations are available, we also report results for the template-based (\texttt{TB}) and facet-based (\texttt{QF-Cond}) baselines of Sekulic et al.~\cite{sekulic_towards_2021}. For LLM-based methods, we use the same underlying model to ensure a fair comparison.

\subsection{Implementation Details and Hyperparameters}

We build on the pre-trained \texttt{LLaMA3.1-8B-base} checkpoint from the HuggingFace Hub, using the \texttt{Transformers} and \texttt{TRL} libraries~\cite{wolf2020transformers}. For supervised fine-tuning (SFT), we train for 2 epochs with a learning rate of $1 \times 10^{-5}$, batch size 32, and a linear learning rate schedule. For direct preference optimization (DPO), we use 2 epochs with a learning rate of $2 \times 10^{-6}$, batch size 32, and $\beta=0.1$. In our joint loss, we set $\gamma=0.5$, based on ablation results. Zero-shot baselines rely on the \texttt{Instruct} variant of the base model. All experiments are run on NVIDIA A100 GPUs (80GB). Source code is available at: \url{https://github.com/RayaneA7/RAC-Retrieval-augmented-clarifcation}.

\section{Results}

\subsection{Main Results}

The main evaluation results are reported in Table~\ref{tab:model_scores_by_dataset}. We find that \(RAC\) significantly outperforms the baselines across all metrics and datasets, confirming that passage conditioning substantially improves clarifying question generation, answering \textbf{RQ2}.

\begin{table}[t]
\centering
\caption{Evaluation scores of RAC variants against different baselines, with $\beta\ =\ 0.1$ and for mixture $\alpha\ =\ 0.7$. Bold values indicate best performance, and $\dagger$ indicates a statistically significant improvement (Welch’s t-test, p < 0.001).}

\label{tab:model_scores_by_dataset}
\resizebox{\textwidth}{!}{%
\begin{tabular}{llcccccc}
\toprule
\textbf{Dataset} & \textbf{Model} & \textbf{ROUGE-L ↑} & \textbf{BLEU ↑} & \textbf{METEOR ↑} & \textbf{BERTScore (F1) ↑} & \textbf{ALScore ↑} & \textbf{Par-R ↑} \\
\midrule
\multicolumn{8}{l}{\textbf{Conversational Search Datasets}} \\
\midrule

\multirow{5}{*}{\textbf{Qulac}} 
& AT-CoT          & 17.97 & 2.77 & 20.81 & 84.72 & -- & -- \\
& Q-Cond          & 29.44 & 10.51 & 25.92 & 88.24 & -- & -- \\
& QP-Zero\textsubscript{shot}          & 27.39 & 5.68 & 33.33 & 87.20 & -- & -- \\
& \textbf{\textbf{RAC\textsubscript{SFT}(ours)}}      & \textbf{33.14}$^{\dagger}$ & \textbf{12.59}$^{\dagger}$ & \textbf{31.30}$^{\dagger}$ & \textbf{89.34}$^{\dagger}$ & 79.14 &  42.53 \\
& + \textbf{RAC\textsubscript{DPO}(ours)}         & \textbf{32.42}$^{\dagger}$ & \textbf{11.52}$^{\dagger}$ & \textbf{31.48}$^{\dagger}$ & \textbf{88.92}$^{\dagger}$ & \textbf{81.73 }& \textbf{44.83} \\

\midrule

\multirow{5}{*}{\textbf{ClariQ}} 
& AT-CoT          & 18.63 & 3.49 & 21.19 & 84.74 & -- & -- \\

& Q-Cond          & 28.68 & 11.19 & 25.47 & 88.16 & -- & -- \\
& TB          & 35.50 & 0.28  & 24.26  & 87.65 & -- & -- \\
& QF-Cond          & 33.70 & 2.20 & \textbf{37.56} & 89.08 & -- & -- \\
& QP-Zero\textsubscript{shot}          & 26.03 & 4.99 & 31.81 & 86.59 & -- & -- \\
& \textbf{\textbf{RAC\textsubscript{SFT}(ours)}}       & \textbf{36.25}$^{\dagger}$ & \textbf{14.88}$^{\dagger}$ & \textbf{34.01}$^{\dagger}$ & \textbf{89.52}$^{\dagger}$ & 51.32 & 53.15 \\
& + \textbf{RAC\textsubscript{DPO}(ours)}    & \textbf{35.52}$^{\dagger}$ & \textbf{14.86}$^{\dagger}$ & \textbf{33.84}$^{\dagger}$ & \textbf{89.39}$^{\dagger}$ & \textbf{52.41} & \textbf{55.77} \\

\midrule
\multicolumn{8}{l}{\textbf{Question Answering Datasets}} \\
\midrule

\multirow{5}{*}{\textbf{PaQa}} 
& AT-CoT          & 23.59 & 7.07 & 22.93 & 85.97 & -- & -- \\
& Q-Cond          & 42.46 & 16.62 & 41.58 & 90.12 & -- & -- \\
    & QP-Zero\textsubscript{shot}         & 33.79 & 10.42 & 35.84 & 88.66 & -- & -- \\
& \textbf{\textbf{RAC\textsubscript{SFT}(ours)}}       & \textbf{46.83}$^{\dagger}$ & \textbf{20.17}$^{\dagger}$ & \textbf{47.97}$^{\dagger}$ & \textbf{90.85}$^{\dagger}$  & 43.36 & 27.62 \\
& + \textbf{RAC\textsubscript{DPO}(ours)}    & \textbf{45.26}$^{\dagger}$ & \textbf{18.32}$^{\dagger}$ & \textbf{46.40}$^{\dagger}$ & \textbf{90.41}$^{\dagger}$ & \textbf{45.75} & \textbf{28.54} \\

\midrule

\multirow{5}{*}{\textbf{CAmbigNQ}} 
& AT-CoT          & 10.33 & 2.10 & 8.53 & 84.02 & -- & -- \\
& Q-Cond          & 28.41 & 8.90 & 33.06 & 87.17     & -- & -- \\
& QP-Zero\textsubscript{shot} & 18.20 & 4.27 & 19.48 & 85.15 & -- & -- \\
& \textbf{RAC\textsubscript{SFT}(ours)}       & \textbf{36.66}$^{\dagger}$ & \textbf{14.81}$^{\dagger}$ & \textbf{43.37}$^{\dagger}$ & \textbf{88.93}$^{\dagger}$ & 47.62 & 87.99 \\
& + \textbf{RAC\textsubscript{DPO}(ours)}   & \textbf{35.47}$^{\dagger}$ & \textbf{14.40}$^{\dagger}$ & \textbf{41.99}$^{\dagger}$ & \textbf{88.89}$^{\dagger}$ & \textbf{49.95} & \textbf{88.05} \\

\bottomrule
\end{tabular}
    }
\end{table}

Moreover, results show that reference-based measures fail to capture the gains from preference tuning, consistent with prior findings~\cite{choi2024model,duong2025scope,rafailov2023direct}. In contrast, reference-free evaluation --reported only for models conditionned with passages-- reveals that \(RAC_{\text{DPO}}\) achieves better performance over \(RAC_{\text{SFT}}\). This demonstrates that preference-based optmization enhances corpus faithfulness beyond sepervised fine-tuning, directly adressing \textbf{RQ3}.

The fact that QP-Zero performs significantly worse than Q-cond highlights the importance of learning the form of a clarification question, independently of its content.

These findings highlight both the benefit of passage conditioning and the added value of preference-based optimization. We further validate these results through qualitative analysis and LLM-based judgments in subsequent experiments.

\subsection{LLM-based Evaluation}

To further address \textbf{RQ2}, we evaluate the faithfulness of our approach using GPT-4 as a evaluator, comparing \(RAC_{\text{DPO}}\) against \(RAC_{\text{SFT}}\). Results are shown in Table \ref{tab:llm-judge}.  
\begin{table}[thb]
\centering
\setlength{\tabcolsep}{8pt}
\renewcommand{\arraystretch}{1.1}
\small
\caption{GPT-4 preference results comparing \(RAC_{\text{DPO}}\) and \(RAC_{\text{SFT}}\). Results with \textbf{*} are statistically significantly different based on the one-sided McNemar’s test with \texttt{p < 0.05}.}
\begin{tabular}{lccc}
\toprule
\textbf{Dataset} & RAC\textsubscript{DPO}\% & Tie\% & RAC\textsubscript{SFT}\% \\
\midrule

Qulac    & \textbf{28.88\textsuperscript{*}} & 50.56 &  20.56 \\
ClariQ   &  \textbf{28.36\textsuperscript{*}} & 48.79 &  22.85 \\
PaQa     &  \textbf{36.24} & 30.36 &  33.40 \\
CAMbigNQ & \textbf{16.27\textsuperscript{*}} & 72.23 &  11.50 \\
\bottomrule
\end{tabular}
\label{tab:llm-judge}
\end{table}
Across all datasets, \(RAC_{\text{DPO}}\) achieves higher win rates compared to \(RAC_{\text{SFT}}\), in some cases by more than a factor of two, whereas a large fraction of outputs are judged as ties. These results suggest that supervised fine-tuning already provides a strong baseline, preference optimization yield further gains on harder cases, reinforcing \textbf{RQ3} by enhancing faithfulness beyond supervised training.
\subsection{Impact of the Number of Input Passages}

We next examine the impact of the number and quality of retrieved passages on RAC. Because RAC relies on retrieval to expose potential ambiguities, both the quantity and relevance of the input passages directly affect its ability to generate effective clarifications.

\begin{figure}[b]
    \centering
    \begin{subfigure}{0.48\linewidth}
        \centering
        \includegraphics[width=\linewidth]{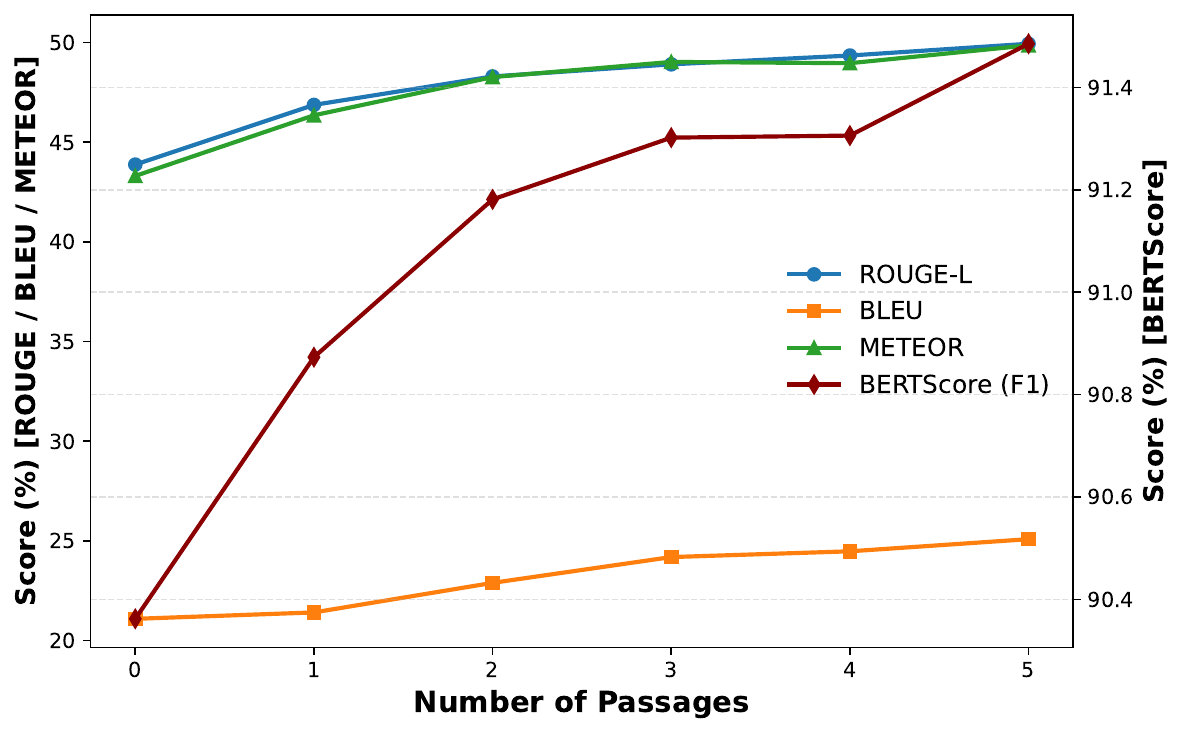}
    \end{subfigure}
    \hfill
    \begin{subfigure}{0.48\linewidth}
        \centering
        \includegraphics[width=\linewidth]{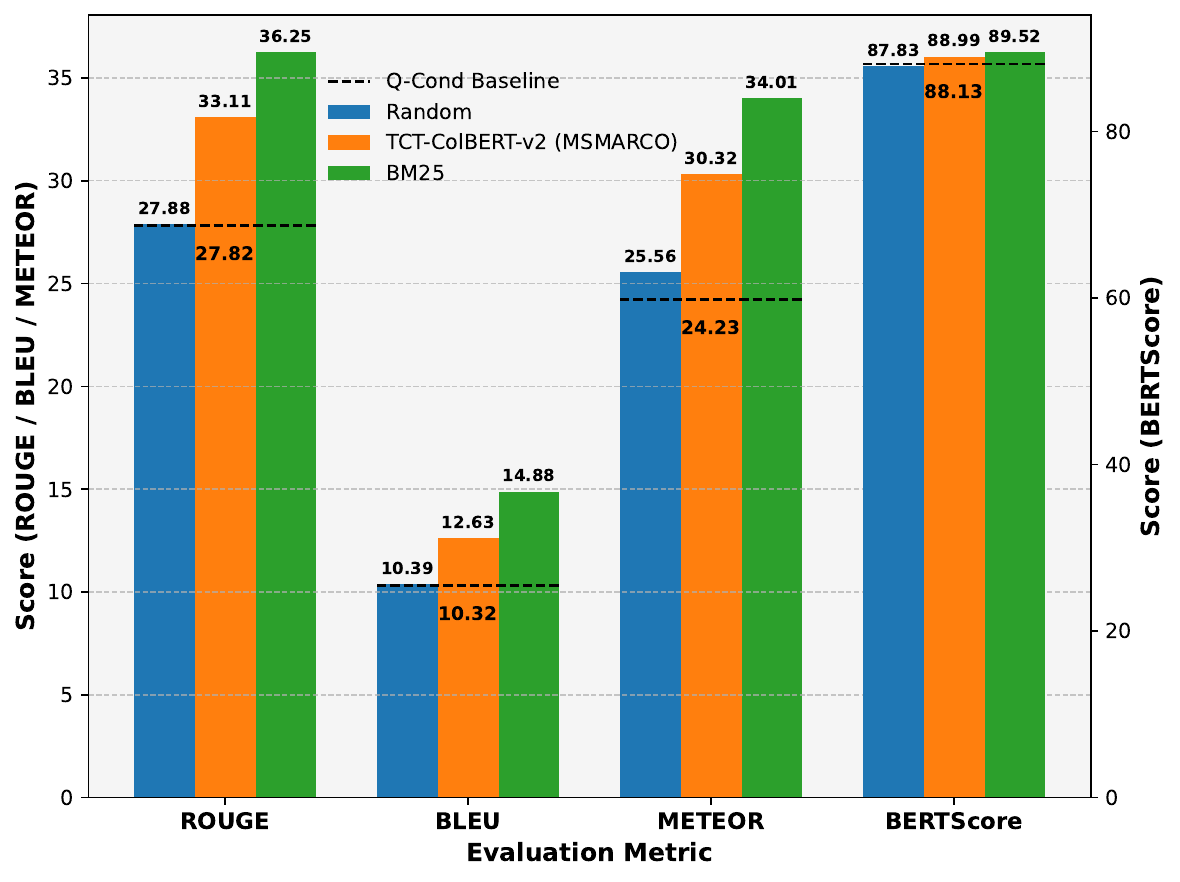}
    \end{subfigure}
\caption{NLG metrics on ClariQ: impact of varying the number of passages (left) and comparison of retrieval strategies (BM25, TCT, random) using the top 5 retrieved passages (right).}

    \label{fig:multi-passage-chart}
\end{figure}

As shown in Fig.~\ref{fig:multi-passage-chart}, performance improves as the number of passages increases, but the effect saturates after approximately four passages, suggesting that the most salient query-related ambiguities are typically captured within the top-ranked results. Passage quality is equally important: using random passages results in performance close to the $"Q\text{-}Cond"$ baseline, whereas BM25 and dense retrievers achieve substantially higher scores. BM25's advantage is likely due to a domain mismatch, since the dense retriever is trained on MS MARCO,
whose passage structure and content differ from the chunked ClueWeb passages used in our setting. These findings indicate that RAC benefits from informative retrieval signals and can extract relevant facets from high-quality passages rather than relying on arbitrary content, thereby addressing \textbf{RQ1}.

\subsection{Impact of the Quality of Noisy Generated Elements}

We study the effect of our noisy generation method by comparing $p_{\text{uncond}}$, fine-tuned with only the query as input, with $p_{\text{LM}}$, the initial language model. We then measure their impact on preference tuning (positive samples always being generated by $p_{\theta_0}$). Table~\ref{tab:negative-generation} shows that $p_{\text{uncond}}$ provides more effective negative samples than $p_{\text{LM}}$. Unlike the approach of Duong et al.~\cite{duong2025scope}, which relies on generic noise injection, $p_{\text{uncond}}$ generates clarifications that are structurally well-formed but factually misaligned. This contrast makes them harder negatives and better training signals for preference optimization. By comparison, samples from $p_{\text{LM}}$ often fail to resemble clarifications at all, limiting their usefulness. These results highlight the importance of tailoring noise generation to the clarification format rather than reusing generic base-model outputs.

\begin{table}[t]
\small
\centering
\caption{ClariQ validation results using different negative generation methods.} 
 \resizebox{\textwidth}{!}{%
\begin{tabular}{lcccccc}
\toprule
\textbf{Method} & \textbf{ROUGE-L} & \textbf{BLEU} & \textbf{METEOR} & \textbf{BERTScore (F1)} & \textbf{ALScore ↑} & \textbf{Par-R ↑} \\
\midrule
$RAC_{DPO},\ C_q^- \sim p_{\text{LM}}$ & 33.84 & 12.93 & 30.79 & 89.25 & 50.81 & 50.73 \\ $RAC_{DPO},\ C_q^- \sim p_{\text{uncond}}$ & \textbf{35.52} & \textbf{14.86} & \textbf{33.84} & \textbf{89.39} & \textbf{52.41} & \textbf{55.77} \\
\hline
\end{tabular}
 }
\label{tab:negative-generation}
\end{table}

\subsection{Qualitative Analysis}

\begin{table}[b]
\centering
\scriptsize
\caption{Noisy generation at different $\alpha$ interpolation weights of $p_{\text{uncond}}$ and $p_{\theta_0}$}

\begin{tabularx}{\textwidth}{l|X}
\toprule
\boldmath$\alpha$ & \textbf{Noisy Generation} \\
\midrule

0.0 & are you interested in the causes of angular cheilitis \\
0.1 & do you want to read an article on that, do you want to be diagnosed \\
0.2 & do you have any recommendations for treating this condition \\
0.3 & do you want to see a \colorbox{red!20}{picture} of angular cheilitis \\
0.4 & would you like to know about symptoms, risk factors, or a way to treat or manage your condition \\
0.5 & would recommend \colorbox{red!20}{a topical or internal medicine} to cure your specific symptom for you based off its symptoms or how do I treat your specific angular cheilitis? \\
0.6 & \colorbox{red!20}{do any methods have been tested?} \\
0.7 & What works to treat this \\
0.8 & Use \colorbox{red!20}{warm soapy water}, \colorbox{red!20}{soft cotton balls}, an \colorbox{red!20}{old razor blade}. If you do not have any of the aforementioned items, you can use a \colorbox{red!20}{regular nail clipper}. \\
0.9 & \colorbox{red!20}{How many times a day should I apply the treatment?} \\

\bottomrule
\end{tabularx}
\label{tab:noisy_generations}
\end{table}

\subsubsection{Noisy clarifying questions.}
We qualitatively assess the effect of mixture between the conditionned \& unconditionned models $p_{\theta_0}$ and \( p_{\text{uncond}} \), controlled by $\alpha$  (Eq.\ref{eq:mix_dist}). At $\alpha=0$, outputs come solely from $p_{\theta_0}$; at $\alpha=1$, from $p_{\text{uncond}}$. Table~\ref{tab:noisy_generations} shows an example from ClariQ, where noise increases with $\alpha$. Irrelevant spans (highlighted in \colorbox{red!20}{red}) illustrate how higher $\alpha$ degrades faithfulness. For preference learning, selecting intermediate $\alpha$ values yields negative examples that are challenging yet informative, avoiding both trivial and overly noisy supervision.

\subsubsection{Generated Clarifying questions.}

We compare clarifications from $RAC_{SFT}$ and $RAC_{DPO}$ on ClariQ validation data (Table~\ref{fig:sft_vs_rac_qualitative}). Faithful content is highlighted in \colorbox{yellow}{yellow}, hallucinations in \colorbox{red!20}{red}. $RAC_{SFT}$ exhibits occasional grounding failures, such as introducing unsupported entities (e.g., hallucinating “season 17”) or omitting relevant evidence (e.g., overlooking “season 16”). In contrast, $RAC_{DPO}$ produces questions more tightly grounded in passages. This aligns with quantitative gains reflecting in more relevant and grounded clarifying questions, demonstrating that DPO improves reliability in ambiguous cases where SFT fails to capture the core ambiguity.
\begin{table}[t]
\centering
\scriptsize
\renewcommand{\arraystretch}{1.5}
\caption{Qualitative comparison of $RAC$ against baseline models.}

\begin{tabular}{|>{\raggedright\arraybackslash}p{6cm}|
                >{\raggedright\arraybackslash}p{3cm}|
                >{\raggedright\arraybackslash}p{3cm}|}
\hline
\textbf{Input} & \textbf{$RAC_{SFT}$} & \textbf{$RAC_{DPO}$} \\
\hline
\textbf{Query:} When does the new family guy season come out?\newline
\textbf{Passages:} ['... Ed O Neill Liam Payne Louis Tomlinson and Neil deGrasse Tyson During this season the guys head to South Korea after', ' Family Guy (season 15) Family Guy season 15 Family Guy fifteenth season premiered on Fox in the United States on September 25 2016 and ended on May 21 2017 The season contained 20 episodes The series follows the dysfunctional Griffin family consisting of father Peter mother Lois daughter Meg son Chris baby ...re Appel and Callaghan Guest stars for the season include Kyle Chandler Stephen Curry Flea Rob Gronkowski']
& \vspace{0pt} 

Are you looking for the release date of Family Guy season \colorbox{yellow}{15, 16} or \colorbox{red!20}{17}?
& \vspace{0pt} 

Are you looking for the release dates of Family Guy seasons \colorbox{yellow}{14, 15 or 16}?\\
\hline
\end{tabular}
\label{fig:sft_vs_rac_qualitative}
\end{table}


\section{Conclusion}

In this work, we introduced clarification question generation as a retrieval conditioned generation task, where questions are generated based on both the user query and retrieved passages. This formulation ensures that clarifications are grounded in information the system can realistically access. Our RAC framework combines retrieval context with preference tuning to improve both the relevance and corpus-faithfulness of generated questions. Experiments on four benchmarks demonstrate that both RAC\textsubscript{SFT} and RAC\textsubscript{DPO} significantly outperform existing baselines, Q-Cond and QP-Zero\textsubscript{shot}, across all reference-based metrics (ROUGE-L, BLEU, METEOR, and BERTScore). We further employ LLM-as-Judge evaluations and novel metrics derived from NLI and data-to-text to quantify the gains in faithfulness to retrieved content of RAC\textsubscript{DPO} over RAC\textsubscript{SFT}, which is critical for conversational search, where the objective is to disambiguate and answer user queries based on retrieved evidence rather than knowledge internal to the language model. As future work, we plan to extend this task to multi-turn clarification and evaluate its impact on downstream retrieval performance.

\begin{credits}
\subsubsection{\ackname} The authors acknowledge the ANR -- FRANCE (French National Research Agency) for its financial support of the GUIDANCE project n$^{\circ}$ANR-23-IAS1-0003 as well as the Chaire Multi-Modal/LLM ANR Cluster IA ANR-23-IACL-0007. This work was granted access to the HPC resources of IDRIS under the allocation AD011016470 made by GENCI.
\subsubsection{\discintname}
The authors have no competing interests to declare that are
relevant to the content of this article. 
\end{credits}

\bibliographystyle{splncs04}  
\bibliography{references}     
\end{document}